\pdfoutput=1

\documentclass[11pt]{article}

\usepackage{acl}

\usepackage{times}
\usepackage{latexsym}
\usepackage{graphicx}
\usepackage{multirow}
\usepackage[inline]{enumitem}

\usepackage[T1]{fontenc}

\usepackage[utf8]{inputenc}

\usepackage{microtype}

\title{Direct Speech-to-Speech Translation With Discrete Units}

\author{Ann Lee$^1$, Peng-Jen Chen$^1$, Changhan Wang$^1$, Jiatao Gu$^1$, Sravya Popuri$^1$, \\
{\bf Xutai Ma$^{1,2}$, Adam Polyak$^1$, Yossi Adi$^1$, Qing He$^1$, Yun Tang$^1$,} \\
{\bf Juan Pino$^1$, Wei-Ning Hsu$^1$} \\
Meta AI$^1$ \quad Johns Hopkins University$^2$ \\
\texttt{\{annl,wnhsu\}@fb.com} \\}

\begin{document}
\maketitle
\begin{abstract}
We present a direct speech-to-speech translation (S2ST) model that translates speech from one language to speech in another language without relying on intermediate text generation.
We tackle the problem by first applying a self-supervised discrete speech encoder on the target speech and then training a sequence-to-sequence speech-to-unit translation (S2UT) model to predict the discrete representations of the target speech.
When target text transcripts are available, we design a joint speech and text training framework that enables the model to generate dual modality output (speech and text) simultaneously in the same inference pass. Experiments on the Fisher Spanish-English dataset show that the proposed framework yields improvement of 6.7 BLEU compared with a baseline direct S2ST model that predicts spectrogram features.
When trained without any text transcripts, our model performance is comparable to models that predict spectrograms and are trained with text supervision, showing the potential of our system for translation between unwritten languages\footnote{Audio samples are available at \url{https://facebookresearch.github.io/speech_translation/direct_s2st_units/index.html}}.
\end{abstract}

\section{Introduction}
\label{sec:intro}
Speech translation aims at converting speech from one language into speech or text in another language. The technology helps bridge the communication barriers between people speaking different languages and can provide access to multimedia content in different languages. Conventional speech-to-text translation (S2T) systems take a cascaded approach by concatenating automatic speech recognition (ASR) and machine translation (MT). In recent years, end-to-end S2T~\citep{berard2016listen} is proposed to alleviate the error propagation issue between ASR and MT. These S2T models can be further combined with text-to-speech (TTS) synthesis to provide both speech and text translation, which allows the technology to be adopted in a wider range of applications.

More recently, researchers have started exploring building direct speech-to-speech translation (S2ST) models without relying on text generation as an intermediate step~\citep{jia2019direct,jia2021translatotron}.
Direct S2ST has the benefits of lower computational costs and inference latency as fewer decoding steps are needed compared to cascaded systems. In addition, direct S2ST is a natural approach for supporting translation for languages without a writing system~\citep{tjandra2019speech,zhang2020uwspeech}.
\citet{jia2019direct} first addresses the problem by training an attention-based sequence-to-sequence model that maps source speech spectrograms into target spectrograms. Model training is challenging as it requires the model to learn not only the alignment between two languages but also the acoustic and linguistic characteristics of both languages.
As a result, there is a performance gap between the direct S2ST system and an S2T+TTS cascaded system.

The recent success in self-supervised learning for speech has demonstrated that speech representations learned from a large unlabelled speech corpus can lead to impressive performance on a variety of downstream tasks~\citep{yang2021superb} including ASR~\citep{baevski2020wav2vec,hsu2021hubert}, speaker and language identification~\citep{fan2020exploring}, etc. Moreover, discretized speech units obtained from the clustering of self-supervised speech representations allow researchers to take advantage of existing NLP modeling techniques on speech, such as spoken generative language modeling~\citep{lakhotia2021generative}.

In this work, we tackle the challenge of modeling target speech in direct S2ST by predicting self-supervised discrete representations of the target speech instead of mel-spectrogram features.
Compared with spectrogram features, self-supervised discrete units can disentangle linguistic content from speaker identify or prosodic information in speech~\citep{polyak2021speech}. With the use of discrete units, we can also apply common practice such as beam search during inference.

We investigate direct S2ST with discrete units in the scenarios where the source and target transcripts may or may not be available, the latter case being representative of unwritten languages. For the written languages, we present a framework that jointly generates speech and text output by combining S2ST and S2T tasks through a shared encoder and a partially shared decoder. We resolve the length mismatch issue between the speech and text output during decoding with connectionist temporal classification (CTC)~\citep{graves2006connectionist}.
Experiments show that with the combination of discrete units prediction, speech and text joint training and beam search, our direct S2ST system matches the performance of a cascaded S2T+TTS system.
For the unwritten target languages, we first extend the use of discrete units to text-to-speech translation~\citep{zhang2020uwspeech} when there are source text transcripts available. Then we show that with multitask learning using both discrete representations for the source and the target speech, it is possible to train a direct S2ST system without the use of any text transcripts.
In addition, we measure the system runtime and memory usage during inference and empirically show that the proposed framework is the most efficient compared to the direct S2ST system that predicts spectrogram features or other cascaded systems.

The rest of this paper is organized as follows. After
introducing background and related work in the next section, we describe our system in detail in Sec.~\ref{sec:model}.
Following this, we present experimental results including objective evaluation on translation quality, subjective evaluation on speech quality, and system benchmark in Sec.~\ref{sec:exp}.
Finally Sec.~\ref{sec:conclusion} concludes with a discussion of potential future work.

\section{Related work}
\label{sec:related}

Conventional S2ST systems are built by combining either cascaded or end-to-end S2T models with TTS~\citep{lavie1997janus,nakamura2006atr}. The majority of the speech translation research has focused on the S2T setup. Studies on ASR+MT systems explore better ways to integrate ASR output lattice to MT models~\citep{matusov2005integration} in order to alleviate the error propagation issue between the two. End-to-end S2T~\citep{berard2016listen} has the potential to resolve the issue, as long as it is properly trained with multitask learning~\citep{weiss2017sequence}, model pre-training~\citep{bahar2019comparative,li2020multilingual} or data augmentation~\citep{jia2019leveraging} to overcome the data scarcity problem. Studies on TTS for S2ST focus more on synthesizing the para-linguistic information transferred from the source speech, such as prosody~\citep{aguero2006prosody,anumanchipalli2012intent} and word-level emphasis~\citep{do2017toward}.

On the other hand, \textit{Translatotron}~\citep{jia2019direct} is an attention-based sequence-to-sequence framework that directly translates mel-spectrogram of the source speech into spectrogram features of the target speech. Multitask learning is essential in facilitating the model to converge, though there is still a performance gap towards a S2T+TTS cascaded system.
The follow-up and concurrent work, \textit{Translatotron 2}~\citep{jia2021translatotron}, addresses the over-generation issue by conditioning the spectrogram synthesizer directly on the output from the auxiliary target phoneme decoder.
\citet{kano2021transformer} propose to build a single deep-learning framework step-by-step by pre-training ASR, MT and TTS models separately and connecting them with Transcoder layers. However, the inference process requires the ASR and MT decoders to complete decoding a full sequence, and thus it loses the latency advantage of a direct S2ST system.
\citet{tjandra2019speech,zhang2020uwspeech} both investigate direct S2ST models under the unwritten language setup by transforming the target speech into discrete representations through Variational Auto-Encoder (VAE), training a sequence-to-sequence model for translation into target discrete units, and an inverter for converting the units to speech. 

In this work, we propose to train a transformer-based speech-to-discrete unit model for direct S2ST. We design a text decoding task conditioned on the intermediate representation of the discrete unit decoder in addition to the auxiliary tasks proposed in~\citep{jia2019direct}.
We choose to use HuBERT~\citep{hsu2021hubert} to generate the target self-supervised discrete units, since~\citet{yang2021superb,lakhotia2021generative,polyak2021speech} have shown its superior performance across ASR, spoken language modeling and speech synthesis, compared to other unsupervised representations, including VAE-based representations used in~\citep{tjandra2019speech,zhang2020uwspeech}.

Overall, there exists little work on direct S2ST due to the lack of parallel S2ST training data. While~\citet{jia2019direct} performs one set of experiments on in-house real-world S2ST data,~\citet{jia2019direct,jia2021translatotron,tjandra2019speech,zhang2020uwspeech,kano2021transformer} all take advantage of TTS services to produce synthetic target speech for model training. We follow the same approach and conduct our experiments with single-speaker synthetic target speech.

\section{Model}
\label{sec:model}
\begin{figure}[t!]
  \centering
  \includegraphics[width=8.5cm]{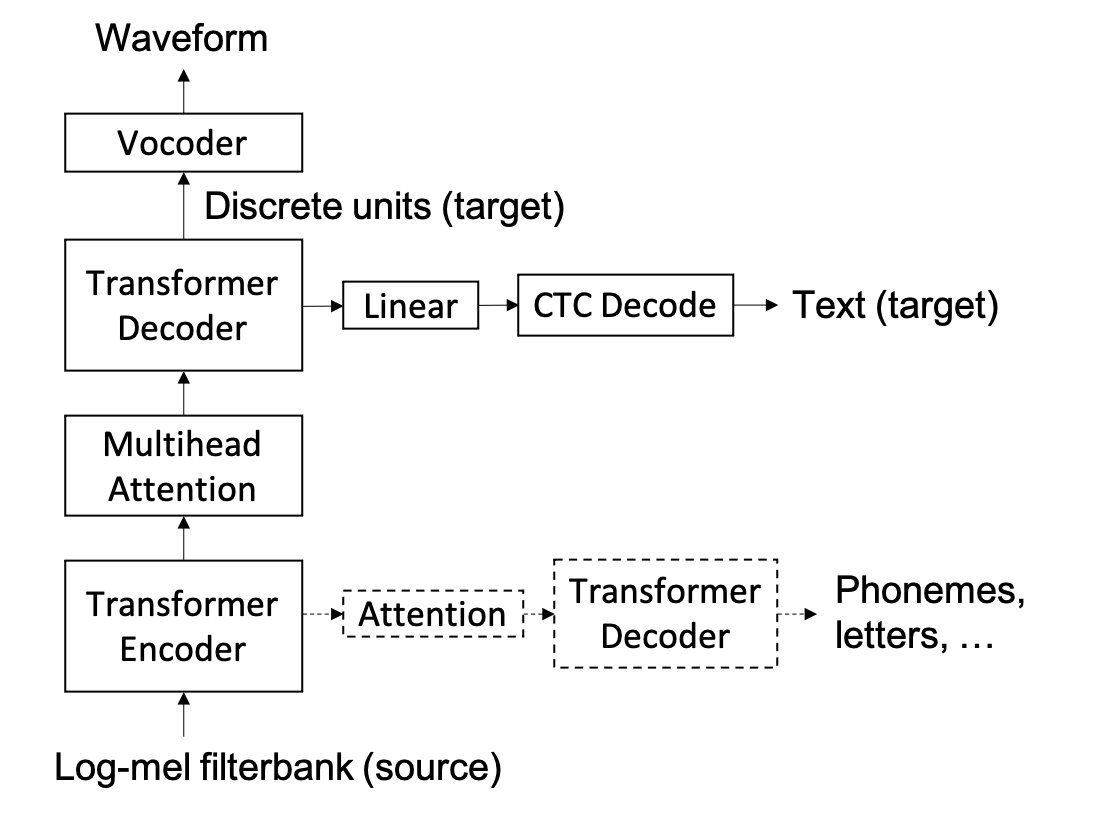}
\caption{An illustration of the direct S2ST model with discrete units. The model consists of
\begin{enumerate*}[label=(\arabic*)]
  \item a transformer-based speech-to-unit translation (S2UT) model with a speech encoder and a discrete unit decoder,
  \item auxiliary tasks conditioned on the encoder,
  \item a text CTC decoder conditioned on the discrete unit decoder, and
  \item a vocoder separately trained to transform discrete units into waveform.
\end{enumerate*}
}
\label{fig:model}
\end{figure}

Our proposed system (Fig.~\ref{fig:model}) is a transformer-based sequence-to-sequence model with a speech encoder and a discrete unit decoder and incorporates auxiliary tasks (shown in dashed lines) similar to~\citep{jia2019direct} during training to facilitate model learning. For written target languages, we further apply target text CTC decoding conditioned on the intermediate representations from the discrete unit decoder for joint speech and text training and generation. Finally, a vocoder is separately trained to convert discrete units into waveform.

\subsection{Speech-to-unit translation (S2UT) model}
HuBERT~\citep{hsu2021hubert} learns speech representations in a self-supervised manner by leveraging k-means clustering on the model's intermediate representations (or the Mel-frequency cepstral coefficient features for the first iteration) to generate discrete labels of masked audio segments.
A HuBERT model pre-trained on an unlabelled speech corpus of the target language can encode the target speech into continuous representations at every 20-ms frame. A k-means algorithm is applied on the learned representations of the unlabelled speech to generate $K$ cluster centroids~\citep{lakhotia2021generative,polyak2021speech}, which are used to encode target utterances into sequences of cluster indices at every 20-ms. In the end, a target utterance $y$ is represented as $[z_1, z_2, ..., z_T], z_i \in \{0, 1, ..., K-1\}, \forall 1 \leq i \leq T$, where $T$ is the number of frames.

\begin{figure}[t!]

\begin{minipage}[b]{.48\linewidth}
  \centering
  \centerline{\includegraphics[height=2.5cm]{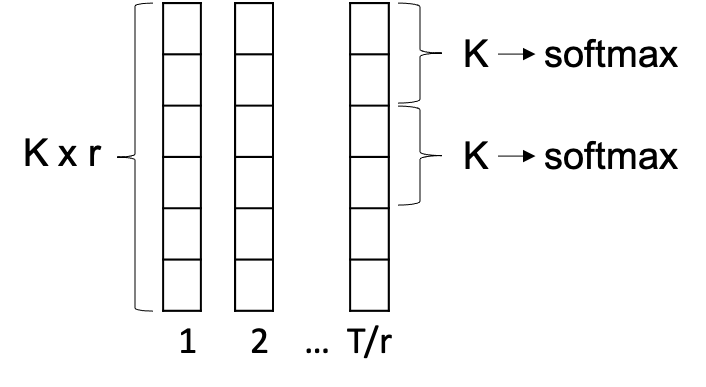}}
  \centerline{(a) \textit{stacked}}\medskip
\end{minipage}
\hfill
\begin{minipage}[b]{0.48\linewidth}
  \centering
  \centerline{\includegraphics[height=2.5cm]{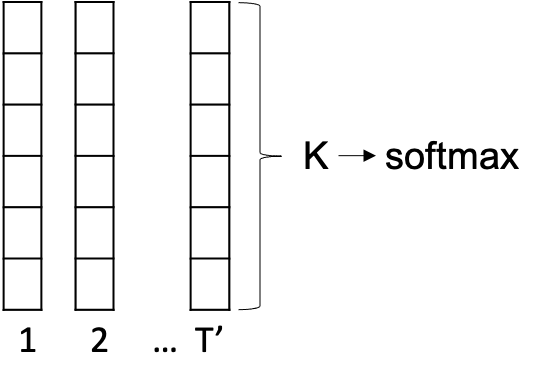}}
  \centerline{(b) \textit{reduced}}\medskip
\end{minipage}
\caption{Two strategies for generating units during decoding. In the \textit{stacked} design ((a)), each decoding step predicts $r$ units by producing a $K \times r$ vector for $r$ softmax computations. In the \textit{reduced} design ((b)), the target unit sequence is reduced to a sequence of unique units with consecutive duplicating units removed.
}
\label{fig:unit}
\end{figure}

We build the S2UT model by adapting from the transformer model for MT~\citep{vaswani2017attention}. A stack of 1D-convolutional layers, each with stride 2 and followed by a gated linear unit activation function, is prepended to the transformer layers in the encoder for downsampling the speech input~\citep{synnaeve2019end}. 
As the target sequence is discrete, we train the S2UT model with cross-entropy loss with label smoothing. We explore two strategies for predicting the discrete unit sequence. In the first strategy (Fig.~\ref{fig:unit}(a), dubbed as ``\textit{stacked}''), we apply the concept of reduction factor, $r$, from TTS~\citep{wang2017tacotron} and generate a $K \times r$ vector at every decoding step for predicting $r$ consecutive discrete units. In the second strategy (Fig.~\ref{fig:unit}(b), dubbed as ``\textit{reduced}''), we collapse a consecutive sequence of the same units into one single unit, resulting a sequence of unique discrete units. Both strategies help speed up training and inference time. 

\subsection{Multitask learning}
We follow the design in~\citep{jia2019direct} to incorporate auxiliary tasks with additional attention and decoder modules conditioned on the intermediate layers of the encoder. The target output of the auxiliary tasks can be either phonemes, characters, subword units or any discrete representations of the source or target utterances. These auxiliary tasks are only used during training and not in inference.

For written target languages, we add target text CTC decoding conditioned on an intermediate layer from the discrete unit decoder for the model to generate dual modality output. The use of CTC can mitigate the length mismatch between the speech and text output. However, since it only allows monotonic alignment, we rely on the transformer layers that the CTC decoder conditioned on to take care of the reordering from source to target. During training, we do teacher-forcing with the ground truth target discrete unit sequence and compute CTC loss using the teacher-forced intermediate representations from the decoder. During inference, we can perform discrete unit decoding and CTC decoding for text at each decode step simultaneously.

\subsection{Unit-based vocoder}
We adopt the modified version of the HiFi-GAN neural vocoder~\citep{kong2020hifi} proposed in~\citep{polyak2021speech} for unit-to-waveform conversion. For the \textit{stacked} discrete unit output, we train the vocoder with only discrete unit sequence and without extra pitch information as the input. For the \textit{reduced} discrete unit output, we enhance the vocoder with a lightweight duration prediction module from Fastspeech 2~\citep{ren2020fastspeech}, which consists of two 1D-convolutional layers, each with ReLU activation and followed by layer normalization and dropout, and a linear layer. We train the enhanced vocoder by minimizing the mean square error (MSE) between the module prediction and the ground truth duration of each unit segment in logarithmic domain, together with the generator-discriminator loss from HiFi-GAN.

\section{Experiments}
\label{sec:exp}

\subsection{Data}
We perform our experiments using the Fisher Spanish-English speech translation corpus~\citep{post2013improveddataset} as in~\citep{jia2019direct,zhang2020uwspeech}. The dataset consists of 139k sentences (approximately 170 hours) from telephone conversations in Spanish, the corresponding Spanish text transcriptions and their English text translation. As in previous studies on direct S2ST~\citep{jia2019direct,jia2021translatotron,zhang2020uwspeech}, we use a high-quality in-house TTS engine to prepare synthetic target speech with a single female voice as the training targets. We perform all experiments, including the baselines, with the synthetic target speech and do not rely on the TTS engine for other uses. We apply the ASR model described in Sec.~\ref{sec:eval} on the synthetic speech and filter out samples with word error rate (WER) greater than 80. Table~\ref{tab:data} lists the statistics of the resulting training set, the two development sets and the test set.

\begin{table}[t]
\centering
\begin{tabular}{c|cccc}
 & train & dev & dev2 & test \\
\hline
\# samples & 126k & 4k & 4k & 3.6k \\
source (hrs) & 162.5 & 4.6 & 4.7 & 4.5 \\
target (hrs) & 139.3 & 4.0 & 3.8 & 3.9 \\
\end{tabular}
\caption{\label{tab:data} Statistics (number of samples and duration) of the Fisher Spanish-English dataset~\citep{post2013improveddataset} after pre-processing }
\end{table}

\subsection{System setup}
\label{sec:system_setup}
\paragraph{S2UT model}
We use the pre-trained HuBERT Base model\footnote{\url{https://github.com/pytorch/fairseq/tree/master/examples/hubert}} trained on Librispeech~\cite{panayotov2015librispeech} for two iterations and follow~\cite{hsu2021hubert,lakhotia2021generative} to perform k-means with $K=100$ on representations from the sixth layer of the model for extracting discrete units for all target English speech. We compute 80-dimensional mel-filterbank features at every 10-ms for the source speech as input to the speech encoder and apply cepstral mean and variance normalization and SpecAugment~\citep{park2019specaugment} with the Librispeech basic policy. The downsampling stack in the speech encoder contains two 1D-convolutional layers with kernel size 5 and 1024 channels, resulting in a downsampling factor of 4 on the input speech. The encoder contains 12 transformer layers with embedding size 256, feed-forward network (FFN) embedding size 2048 and 4 attention heads. The decoder consists of 6 transformer layers with the same embedding size and FFN embedding size as the encoder and 8 attention heads. 

We explore four targets for the auxiliary tasks: source phonemes (\textit{sp}), target phonemes (\textit{tp}), source characters (\textit{sc}) and target characters (\textit{tc}).
For \textit{sp} or \textit{sc}, we append an attention module and a decoder to the sixth layer of the encoder based on preliminary experimentation.
For \textit{tp} or \textit{tc}, we attach the attention and the decoder to the eighth layer of the encoder.
All multihead attention modules have 4 heads and the decoders have 2 transformer layers, 256-dimensional embeddings and a FFN embedding size of 2048. Each auxiliary loss has a constant weight of 8.0 during training.

For written target languages, we condition the CTC decoding on the third layer of the discrete unit decoder. The target text for CTC is encoded as 1k unigram subword units~\cite{kudo2018subword} to guarantee that the text sequence length is shorter than the length of the \textit{stacked} or \textit{reduced} discrete unit sequence. The weight on the CTC loss is set to 1.6 during training. We train the models for 400k steps using Adam with $\beta_{1}=0.9, \beta_{2}=0.98, \epsilon=10^{-8}$, label smoothing $0.2$, and apply an inverse square root learning rate decay schedule with 10k warmup steps. All other hyper-parameters, such as dropout and learning rate, are tuned on the development set.
All models are implemented using \textsc{fairseq} S2T~\cite{ott2019fairseq,wang2020fairseqs2t}\footnote{Code is available at \url{https://github.com/pytorch/fairseq/tree/main/examples/speech_to_speech}.}.

\paragraph{Unit-based vocoder}
We follow the same vocoder design and training procedure in~\citep{polyak2021speech} and incorporate a duration prediction module from Fastspeech 2~\citep{ren2020fastspeech}. The two 1D-convolutional layers in the module have a filter size of 128 and a kernel size of 3. We apply a dropout of 0.5, and the weight on the MSE loss from the duration prediction module is set to 1.0 during training\footnote{Code for vocoder training is available at \url{https://github.com/facebookresearch/speech-resynthesis/tree/main/examples/speech_to_speech_translation}}. The vocoder is trained on the synthetic target speech from the Fisher training set.

\begin{table*}[t]
\centering
\resizebox{.99\linewidth}{!}{
\begin{tabular}{c|lcccccc|c}
\hline
 & & \multicolumn{6}{c|}{BLEU} & MOS \\
 & & \multicolumn{2}{c}{dev} & \multicolumn{2}{c}{dev2} & \multicolumn{2}{c|}{test} & test \\
ID & & speech & text & speech & text & speech & text & \\
\hline
1 & Synthetic target  & 88.5 & 100.0 & 89.4 & 100.0 & 90.5 & 100.0 & 3.49 $\pm$ 0.14 \\
\hline
& Cascaded systems: \\
2 & ASR (beam=10) + MT (beam=5) + TTS & 42.1 & 45.1 & 43.5 & 46.1 & 43.9& 46.3 & 3.37 $\pm$ 0.15 \\
3 & S2T (beam=10) + TTS & 38.5 & 41.1 & 39.9 & 42.4 & 40.2 & 42.1 & 3.43 $\pm$ 0.14 \\
\hline
& Direct systems: \\
4 & Transformer \textit{Translatotron} ($r=5$, w/ \textit{sp, tp}) & 25.0 & - & 26.3 & - & 26.2 & - & - \\
5 & Transformer \textit{Translatotron} ($r=5$, w/ \textit{sc, tc}) & 32.9 & - & 34.1 & - & 33.2 & - & 3.31 $\pm$ 0.11 \\
6 & S2UT, no reduction ($r=1$, w/ \textit{sc, tc}) & 33.4 & - & 34.6 & - & 34.1 & - & 3.35 $\pm$ 0.14 \\
7 & S2UT \textit{stacked} ($r=5$, w/ \textit{sc, tc}) & 34.0 & - & 34.5 & - & 34.4 & - & - \\
\hline
& Direct systems with dual modality output: \\
8 & S2UT \textit{stacked} + CTC ($r=5$, w/ \textit{sc, tc}) & 34.4 & 36.4 & 36.4 & 37.9 & 34.4 & 35.8 & 3.32 $\pm$ 0.14 \\ 
9 & S2UT \textit{reduced} + CTC (w/ \textit{sc, tc}), beam=1 & 36.8 & 40.0 & 38.4 & 41.5 & 38.5 & 40.7 & - \\
10 & S2UT \textit{reduced} + CTC (w/ \textit{sc, tc}), beam=10 & \textbf{38.2} & \textbf{41.3} & \textbf{39.5} & \textbf{42.2} & \textbf{39.9} & \textbf{41.9} & 3.41 $\pm$ 0.14 \\
\hline
\hline
& From the literature$^\ast$: \\
11 & \textit{Translatotron}~\cite{jia2019direct} & 24.8 & - & 26.5 & - & 25.6 & - & 3.69 $\pm$ 0.07 \\
12 & + pre-trained encoder~\cite{jia2019direct} & 30.1 & - & 31.5 & - & 31.1 & - & - \\
13 & \textit{Translatotron 2}~\cite{jia2021translatotron} & - & - & - & - & 37.0 & - & 3.98 $\pm$ 0.08 \\
14 & + data augmentation~\cite{jia2021translatotron} & - & - & - & - & 40.3 & - & 3.79 $\pm$ 0.09 \\
\hline
\end{tabular}
}
\caption{\label{tab:written} Results from systems using target transcripts during training. Translation content quality is evaluated via BLEU scores with respect to four references from the Fisher Spanish-English dataset. For systems generating dual modality output (cascaded and S2UT + CTC), we evaluate both the text output directly from the system and the ASR decoded text from the speech output. We only evaluate the latter for systems generating speech-only output. Speech naturalness is evaluated via a subjective MOS test, and we report MOS results with 95\% confidence interval. ($^\ast$: results are not directly comparable due to different ASR models and MOS protocols used for evaluation.)}
\end{table*}

\subsection{Baselines}
\label{sec:baseline}
We build two cascaded baselines, ASR+MT+TTS and S2T+TTS, and one direct S2ST baseline that predicts spectrogram features. All models in the cascaded baselines are trained with character input or output.
\paragraph{ASR}
We train the transformer-based Spanish ASR system with the default hyper-parameters and \texttt{s2t\_transformer\_s} architecture in \textsc{fairseq} S2T~\citep{wang2020fairseqs2t}.
\paragraph{MT}
As the input to the MT model is characters, we follow the default \texttt{gru\_transformer} setup in \textsc{fairseq}~\citep{ott2019fairseq} to prepend a bidirectional recurrent layer with gated recurrent units (GRU) to the transformer encoder to incorporate a larger context~\cite{wang2020neural}.
\paragraph{S2T}
We explore both LSTM-based~\citep{weiss2017sequence} and transformer-based end-to-end S2T systems. The former consists of 8 bidirectional LSTM layers for the encoder and 4 LSTM layers for the decoder. Embedding and hidden state sizes are all 256. The latter has the same model architecture as the S2UT model except that it predicts characters as output. We do not apply pre-training or multitask learning and find that the LSTM-based model works better.

\paragraph{TTS}
The transformer-based TTS model~\citep{li2019neural} has 6 transformer layers, 4 attention heads, embedding size 512 and FFN embedding size 2048 for both the encoder and the decoder. We use 32-dimensional layer for the decoder prenet. The model is trained on the English text and the synthetic target speech with a reduction factor of 5 on the output feature frames. The vocoder is a HiFi-GAN model~\citep{kong2020hifi} fine-tuned on the mel-spectrogram features from teacher-forcing.

\paragraph{Transformer \textit{Translatotron}}
We implement a transformer-based \textit{Translatotron} instead of the LSTM architecture in~\citep{jia2019direct} to speed up model training. The model predicts mel-spectrogram features of the target speech and consists of the same speech encoder design as in the S2UT model, the same speech decoder design as in the TTS model for the cascaded baselines, and a fine-tuned HiFi-GAN vocoder~\citep{kong2020hifi}.
We use the same auxiliary task setup as in the S2UT model with a constant weight of 0.1 on each auxiliary loss, apply a reduction factor of 5 on the output feature frames and tune the hyper-parameters on the development sets. Preliminary studies show no performance degradation for the transformer-based model compared with our implementation of the LSTM version of the model.

\subsection{Evaluation}
\label{sec:eval}
We evaluate both the translation quality and the speech quality of the system output. To evaluate the translation quality, we follow the setup in~\cite{jia2019direct,zhang2020uwspeech} to apply ASR on the speech output and compute BLEU scores of the ASR decoded text with respect to the reference translations.
We adopt an open-sourced English ASR model\footnote{\url{https://github.com/pytorch/fairseq/tree/master/examples/wav2vec}} built with the combination of wav2vec 2.0 pre-training and self-training~\citep{xu2021self}. The model, which is pre-trained on Libri-Light~\citep{kahn2020libri} and fine-tuned on full Librispeech~\citep{panayotov2015librispeech}, achieves WER of 1.9 and 3.9 on the Librispeech test-clean and other sets, respectively. As the ASR output is in lowercase and without punctuation except apostrophes, we normalize the reference text before computing BLEU using \textsc{sacreBLEU}~\citep{post2018call}\footnote{\textsc{sacreBLEU} signature: nrefs:4|case:lc|eff:no|tok:13a|smooth:exp|version:2.0.0}.

In addition to measuring the translation quality via an objective metric, we conduct human listening tests to collect mean opinion scores (MOS) to evaluate the naturalness of the speech output. We randomly sample 200 utterances from the test set, and each sample is rated by 8 raters on a scale of 1–5, with 1 being the worst and 5 being the best.

\subsection{Results}
We explore model training under both written and unwritten language scenarios.
For the former, we take advantage of text transcriptions of source and target speech during S2UT model training. For the latter, we focus on the cases where the source is in either a written or unwritten language, while the target language is without a writing system.

\paragraph{Source \& Target Written}
Table~\ref{tab:written} summarizes the experimental results under the written language setup.
In the following discussion, we first focus on the translation content quality evaluated by BLEU. 
We include the results from~\cite{jia2019direct,jia2021translatotron} as references (11-14). However, as different ASR models are used for evaluation, we should not directly compare the BLEU scores with our experiments. We also list the BLEU scores evaluated on the synthetic target speech (1) to show the impact of the ASR errors on the evaluation metric.

First, we explore using different targets for the auxiliary tasks with transformer \textit{Translatotron} and see that using characters as targets for the auxiliary tasks gives 7 BLEU gain on the test set compared to phonemes (4 vs.~5). In all following experiments, we use characters as the auxiliary task targets.

Second, we compare the proposed S2UT model to transformer \textit{Translatotron}. We start with the \textit{stacked} strategy as both models have the same reduction ratio of 5. We can see that ``S2UT \textit{stacked}'' outperforms the transformer \textit{Translatotron} by 1.2 BLEU on the test set (5 vs.~7), indicating that discrete units are easier to model than continuous-valued mel-spectrogram features. We also experiment with S2UT training using the full discrete unit sequence ($r=1$) and see that a larger reduction factor can speed up training and inference and does not hurt the performance (6 vs.~7).

Third, we incorporate target text CTC decoding to the S2UT model and evaluate both speech and text output. Joint training with discrete unit loss and text CTC loss brings an average gain of 1.2 BLEU on the dev sets for ``S2UT \textit{stacked}'' (7 vs.~8), while the performance on the test set remains the same. Moreover, we see that the \textit{reduced} strategy is more effective than \textit{stacked}. When decoding with a beam size of 1, we see 1.4 BLEU improvement on speech output and 1.2 BLEU gain on text output on the test set (8 vs.~9).

Finally, we apply beam search on the best setup we find, ``S2UT \textit{reduced}'' with joint speech and text training and auxiliary tasks, and the resulting direct S2ST system performs on par with the S2T+TTS system (3 vs.~10) and bridges 63\% of the gap between transformer \textit{Translatotron} (5) and the three-stage ASR+MT+TTS cascaded system (2).
Compared with the cascaded system, the proposed framework has the advantage of being able to generate consistent speech and text output in one inference pass.
We also examine the output from the \textit{tc} auxiliary task, which can serve as another way to generate translated text from the direct S2ST system. By using ASR decoded text from the speech output as reference, we see a character error rate (CER) of 4.5 for the CTC decoded text and 30.3 for the \textit{tc} decoded text on the dev set, indicating that the former is more aligned with the generated audio.

From the MOS results in Table~\ref{tab:written}, we see that direct S2ST systems that predict all frames, such as \textit{Translatotron} and S2UT \textit{stacked} models, tend to have slightly lower MOS than others. The proposed S2UT \textit{reduced} system has an MOS close to that for synthetic target (1 vs.~10). The latter can be viewed as the upper bound of the best MOS we can get, since the model is trained with the synthetic speech as target.

\begin{table}[t]
\centering
\resizebox{.98\linewidth}{!}{
\begin{tabular}{c|lccc}
\hline
ID & BLEU & dev & dev2 & test \\
\hline
& \multicolumn{4}{l}{\textbf{source written}} \\
& Cascaded systems: \\
15 & ASR + T2ST ($r=2$) & 25.3 & 25.5 & 25.9 \\
16 & ASR + T2UT \textit{reduced} & 39.9 & 40.6 & 41.0 \\
\hline
& Direct system: \\
17 & S2UT \textit{reduced} (w/ \textit{sc}) & 34.4 & 35.4 & 35.2\\
\hline
& From the literature$^\ast$: \\
18 & \textit{Translatotron} (w/ \textit{sp})~\citep{jia2019direct} & 7.4 & 8.0 & 7.2 \\
\hline
\hline
& \multicolumn{4}{l}{\textbf{source unwritten}} \\
& Direct systems: \\
19 & S2UT \textit{reduced}, no auxiliary task & 7.8 & 8.0 & 7.4 \\
20 & S2UT \textit{reduced} (w/ \textit{su}) & 31.1 & 32.2 & 31.8 \\
\hline
& From the literature$^\ast$: \\
\multirow{2}{*}{21} & \textit{Translatotron}, no auxiliary task & \multirow{2}{*}{0.4} & \multirow{2}{*}{0.6} & \multirow{2}{*}{0.6} \\
& \citep{jia2019direct} \\
22 & \textit{UWSpeech~\citep{zhang2020uwspeech}} & - & - & 9.4 \\
\hline
\end{tabular}}
\caption{\label{tab:unwritten}  Results from systems trained without using any target text transcripts. BLEU scores are evaluated on the ASR decoded text of the speech output with respect to four references from the Fisher Spanish-English dataset. We use beam size 10 when decoding all S2UT systems. ($^\ast$: results are not directly comparable due to different ASR models used for evaluation.)
}
\end{table}

\begin{figure*}[t!]

\begin{minipage}[b]{.32\linewidth}
  \centering
  \centerline{\includegraphics[width=5.2cm]{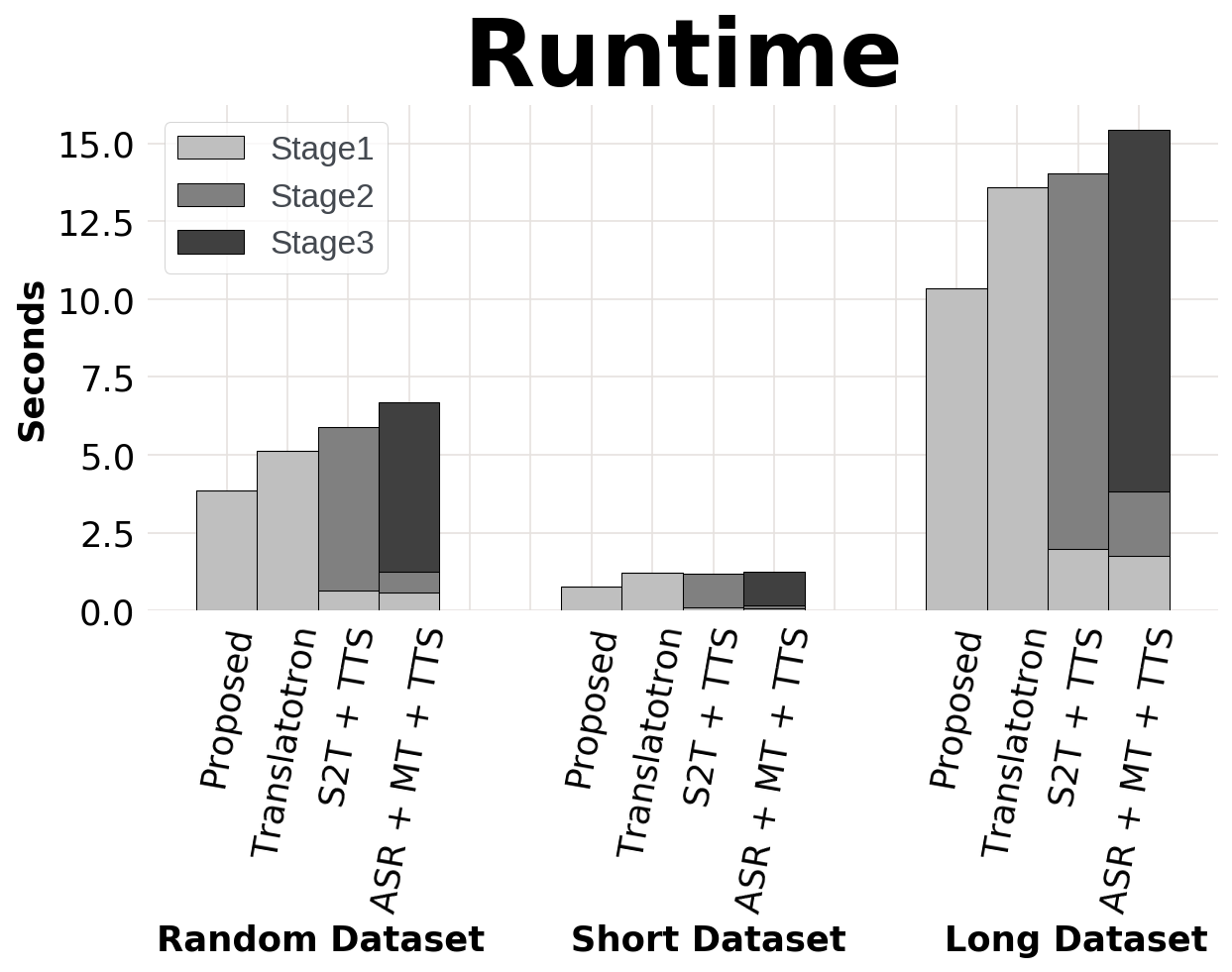}}
  \centerline{(a) \textit{Runtime in seconds}}\medskip
\end{minipage}
\hfill
\begin{minipage}[b]{0.32\linewidth}
  \centering
  \centerline{\includegraphics[width=5.2cm]{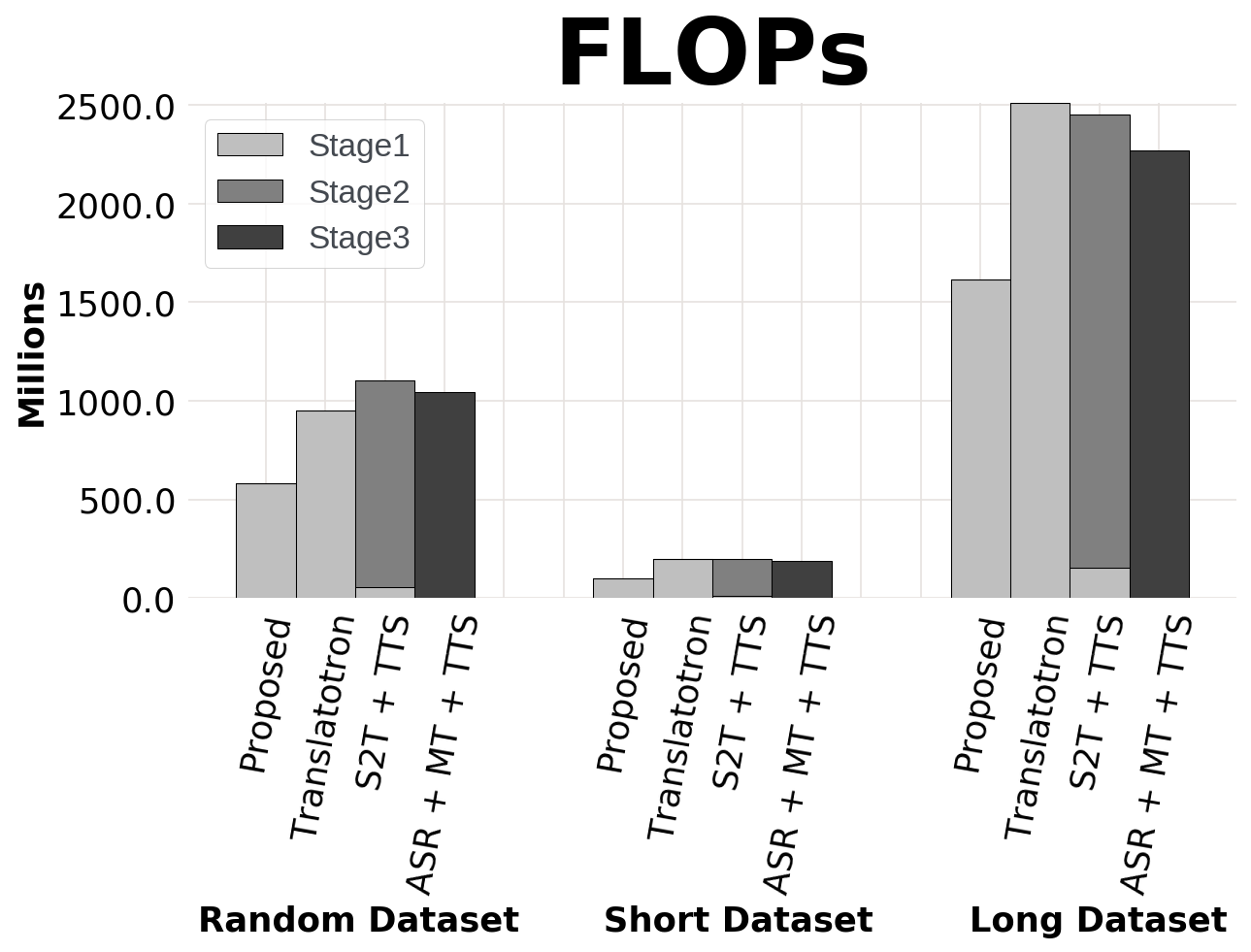}}
  \centerline{(b) \textit{FLOPs in Millions}}\medskip
\end{minipage}
\hfill
\begin{minipage}[b]{0.32\linewidth}
  \centering
  \centerline{\includegraphics[width=5.2cm]{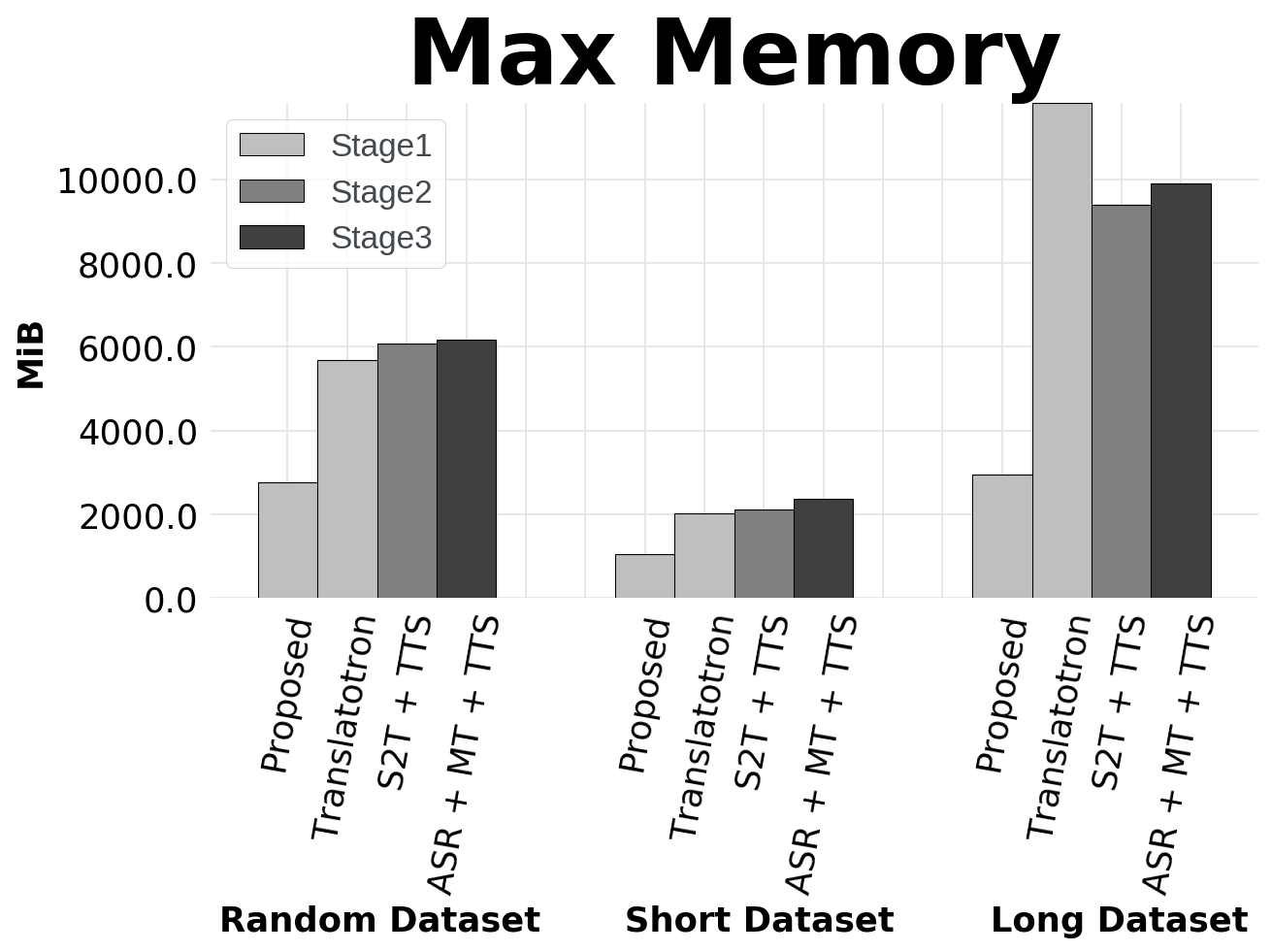}}
  \centerline{(c) \textit{Max Memory in MiB}}\medskip
\end{minipage}
\caption{Benchmarking results for two direct S2ST systems (proposed S2UT \textit{reduced} and transformer \textit{Translatotron}), one two-stage (S2T+TTS), and one three-stage cascaded system (ASR+MT+TTS). We examine runtime, FLOPs and max memory usage on three subsets sampled randomly or from either the shortest or the longest samples from the Fisher dev set.}
\label{fig:benchmark}
\end{figure*}

\paragraph{Source Written, Target Unwritten}
We explore the unwritten target language setup by starting from the scenario where the source speech has a text writing system.
Table~\ref{tab:unwritten} summarizes the results.

First, we build cascaded systems by combining ASR and text-to-speech translation~\citep{zhang2020uwspeech}. The latter can be built by either training a TTS model that predicts spectrogram features or a text-to-unit model with source text and target speech in two languages. We refer to the first approach as text-to-spectrogram translation (T2ST) and the second as text-to-unit translation (T2UT).
We use the same architecture as the transformer TTS model to train the T2ST model with reduction ratio 2, and the same setup as the MT model to train the T2UT model with \textit{reduced} unit sequences.
From Table~\ref{tab:unwritten}, we see that the model that predicts discrete units outperforms the one that predicts spectrogram features by 15.1 BLEU on the test set (15 vs.~16), which is another evidence showing that discrete units are easier to model as translation targets than continuous spectrogram features.
In fact, ASR+T2UT also outperforms S2T+TTS by 0.8 BLEU on the test set (3 vs.~16), which provides another option for building two-stage cascaded systems.

Next, we focus on ``S2UT \textit{reduced}'' based on the findings from the written language setup for direct S2ST. We find that training an S2UT model with \textit{sc} auxiliary task can already achieve 88\% of the performance from a system trained with both source and target text (10 vs.~17).
This is in contrary to the findings in~\cite{jia2019direct} where training \textit{Translatotron} with only source transcripts attains 28\% of the performance of a system trained with both source and target text.

\paragraph{Source \& Target Unwritten}
We extend our experiments to a fully unwritten language setup by training models without using any text transcripts (Table~\ref{tab:unwritten}).~\citet{jia2019direct} has pointed out that the model has difficulty in learning to attend to the input speech when trained without auxiliary tasks.~\citet{zhang2020uwspeech} addresses the challenge by training with discrete unit targets and shows potential, while it uses labelled speech from languages other than the source or the target to guide the VAE learning for the discrete units.

When ``S2UT \textit{reduced}'' is trained without auxiliary tasks, the performance greatly deteriorates (19). We notice that the model can still generate meaningful text. However, the generated speech does not reflect the content in the source speech, and the 7.4 BLEU score is mostly contributed by the function words. This shows that the discrete unit decoder can learn a language model over the unit sequence, while the challenge is in the attention on the encoder output.

To facilitate the S2UT model training, we apply the HuBERT model pre-trained on English to extract discrete representations for the source Spanish speech, and the source units (\textit{su}) are used as an auxiliary task target. The resulting S2UT model achieves only a 1.4 BLEU difference on the test set compared with transformer \textit{Translatotron} trained with both source and target text supervision (5 vs.~20). This shows that source units are effective in guiding the model to properly learn the attention, and the self-supervised discrete representations can capture basic pronunciations that are transferable across languages.

\subsection{System benchmark}
In addition to evaluating the quality of the system output, we examine the efficiency of the models during inference by benchmarking the runtime, total number of floating point operations (FLOPs) and max memory on an Intel\textregistered\ Xeon\textregistered\ Gold 6230 CPU. We conduct the study with three subsets of 500 samples from the Fisher dev set, one with random samples, one with the shortest and the other one with the longest utterances.

Fig.~\ref{fig:benchmark} shows the comparison of two direct S2ST systems, the proposed S2UT \textit{reduced} and transformer \textit{Translatotron}, one two-stage cascaded system (S2T+TTS) and one three-stage cascaded system (ASR+MT+TTS).
For each system, we report the runtime and FLOPs measured by \texttt{timeit} and \texttt{PyPAPI} from all stages, and the maximum memory from any single stage measured by \texttt{memory-profiler}. All metrics are averaged by the total number of samples. For cascaded models we only consider the metrics for model inference at different stages and ignore any intermediate data/IO processing overhead.

First, we see that TTS is the bottleneck for cascaded systems, as it takes up the largest percentage of runtime ($>$89\% in S2T+TTS and $>$81\% in ASR+MT+TTS) and contributes to the maximum memory used. The runtime may be improved with the use of non-autoregressive TTS systems. We leave the investigation to future work, as it is also possible to apply non-autoregressive translation with discrete units.

Next, the proposed S2UT \textit{reduced} model is the most efficient among the four systems across all subsets. Compared to S2T+TTS, our direct system runs 1.5X faster and reduces 47\% FLOPs and 55\% max memory, while maintaining the same level of translation quality (Table~\ref{tab:written}). This verifies one of the benefits of direct S2ST systems, which is lower computational costs and inference latency.

Lastly, the proposed S2UT \textit{reduced} can not only produce better translation than transformer \textit{Translatotron} but also run 1.3X faster and reduce 39\% FLOPs and 51\% max memory. This demonstrates an addition advantage of modeling discrete units instead of spectrogram features.

\section{Conclusion}
\label{sec:conclusion}
We investigate training direct S2ST models with the use of self-supervised discrete representations as targets. We examine model training under both the written and unwritten language scenarios. For the former, we propose a framework with joint speech and text training that performs on par with an S2T+TTS baseline, yet it can run more efficiently. We demonstrate the possibility of translating between two unwritten languages by taking advantage of discrete representations of both the source and the target speech for model training.
Our empirical benchmark shows that the proposed direct S2ST system with discrete units is the most efficient during inference compared with a direct S2ST model that predicts spectrogram features or other cascaded systems.

With the recent release of large-scale S2S dataset~\cite{wang2021voxpopuli}, we plan to investigate the proposed framework with real data in the future.
Another important aspect in generating speech output is the voice and prosody. In our work, we focus on content translation and leave the para-linguistic aspect of speech translation to future work.

We use an open-sourced ASR model for evaluation, so the results should be comparable with all future research in the field that follows the same evaluation protocol. We will also release the code for reproducing the experiments.

\section*{Acknowledgement}
We would like to thank Jade Copet, Emmanuel Dupoux, Evgeny Kharitonov, Kushal Lakhotia, Abdelrahman Mohamed, Tu Anh Nguyen and Morgane Rivière for helpful discussions on discrete representations.

\bibliography{main_acl}

\begin{thebibliography}{40}
\expandafter\ifx\csname natexlab\endcsname\relax\def\natexlab#1{#1}\fi

\bibitem[{Aguero et~al.(2006)Aguero, Adell, and Bonafonte}]{aguero2006prosody}
PD~Aguero, Jordi Adell, and Antonio Bonafonte. 2006.
\newblock Prosody generation for speech-to-speech translation.
\newblock In \emph{2006 IEEE International Conference on Acoustics Speech and
  Signal Processing Proceedings}, volume~1, pages I--I. IEEE.

\bibitem[{Anumanchipalli et~al.(2012)Anumanchipalli, Oliveira, and
  Black}]{anumanchipalli2012intent}
Gopala~Krishna Anumanchipalli, Luis~C Oliveira, and Alan~W Black. 2012.
\newblock Intent transfer in speech-to-speech machine translation.
\newblock In \emph{2012 IEEE Spoken Language Technology Workshop (SLT)}, pages
  153--158. IEEE.

\bibitem[{Baevski et~al.(2020)Baevski, Zhou, Mohamed, and
  Auli}]{baevski2020wav2vec}
Alexei Baevski, Yuhao Zhou, Abdelrahman Mohamed, and Michael Auli. 2020.
\newblock wav2vec 2.0: A framework for self-supervised learning of speech
  representations.
\newblock \emph{Advances in Neural Information Processing Systems}, 33.

\bibitem[{Bahar et~al.(2019)Bahar, Bieschke, and Ney}]{bahar2019comparative}
Parnia Bahar, Tobias Bieschke, and Hermann Ney. 2019.
\newblock A comparative study on end-to-end speech to text translation.
\newblock In \emph{2019 IEEE Automatic Speech Recognition and Understanding
  Workshop (ASRU)}, pages 792--799. IEEE.

\bibitem[{B{\'e}rard et~al.(2016)B{\'e}rard, Pietquin, Servan, and
  Besacier}]{berard2016listen}
Alexandre B{\'e}rard, Olivier Pietquin, Christophe Servan, and Laurent
  Besacier. 2016.
\newblock Listen and translate: A proof of concept for end-to-end
  speech-to-text translation.
\newblock \emph{arXiv preprint arXiv:1612.01744}.

\bibitem[{Do et~al.(2017)Do, Sakti, and Nakamura}]{do2017toward}
Quoc~Truong Do, Sakriani Sakti, and Satoshi Nakamura. 2017.
\newblock Toward expressive speech translation: A unified sequence-to-sequence
  {LSTM}s approach for translating words and emphasis.
\newblock In \emph{INTERSPEECH}, pages 2640--2644.

\bibitem[{Fan et~al.(2020)Fan, Li, Zhou, and Xu}]{fan2020exploring}
Zhiyun Fan, Meng Li, Shiyu Zhou, and Bo~Xu. 2020.
\newblock Exploring wav2vec 2.0 on speaker verification and language
  identification.
\newblock \emph{arXiv preprint arXiv:2012.06185}.

\bibitem[{Graves et~al.(2006)Graves, Fern{\'a}ndez, Gomez, and
  Schmidhuber}]{graves2006connectionist}
Alex Graves, Santiago Fern{\'a}ndez, Faustino Gomez, and J{\"u}rgen
  Schmidhuber. 2006.
\newblock Connectionist temporal classification: labelling unsegmented sequence
  data with recurrent neural networks.
\newblock In \emph{Proceedings of the 23rd international conference on Machine
  learning}, pages 369--376.

\bibitem[{Hsu et~al.(2021)Hsu, Bolte, Tsai, Lakhotia, Salakhutdinov, and
  Mohamed}]{hsu2021hubert}
Wei-Ning Hsu, Benjamin Bolte, Yao-Hung~Hubert Tsai, Kushal Lakhotia, Ruslan
  Salakhutdinov, and Abdelrahman Mohamed. 2021.
\newblock {H}u{BERT}: Self-supervised speech representation learning by masked
  prediction of hidden units.
\newblock \emph{arXiv preprint arXiv:2106.07447}.

\bibitem[{Jia et~al.(2019{\natexlab{a}})Jia, Johnson, Macherey, Weiss, Cao,
  Chiu, Ari, Laurenzo, and Wu}]{jia2019leveraging}
Ye~Jia, Melvin Johnson, Wolfgang Macherey, Ron~J Weiss, Yuan Cao, Chung-Cheng
  Chiu, Naveen Ari, Stella Laurenzo, and Yonghui Wu. 2019{\natexlab{a}}.
\newblock Leveraging weakly supervised data to improve end-to-end
  speech-to-text translation.
\newblock In \emph{ICASSP 2019-2019 IEEE International Conference on Acoustics,
  Speech and Signal Processing (ICASSP)}, pages 7180--7184. IEEE.

\bibitem[{Jia et~al.(2021)Jia, Ramanovich, Remez, and
  Pomerantz}]{jia2021translatotron}
Ye~Jia, Michelle~Tadmor Ramanovich, Tal Remez, and Roi Pomerantz. 2021.
\newblock Translatotron 2: Robust direct speech-to-speech translation.
\newblock \emph{arXiv preprint arXiv:2107.08661}.

\bibitem[{Jia et~al.(2019{\natexlab{b}})Jia, Weiss, Biadsy, Macherey, Johnson,
  Chen, and Wu}]{jia2019direct}
Ye~Jia, Ron~J Weiss, Fadi Biadsy, Wolfgang Macherey, Melvin Johnson, Zhifeng
  Chen, and Yonghui Wu. 2019{\natexlab{b}}.
\newblock Direct speech-to-speech translation with a sequence-to-sequence
  model.
\newblock \emph{Proc. Interspeech 2019}, pages 1123--1127.

\bibitem[{Kahn et~al.(2020)Kahn, Rivi{\`e}re, Zheng, Kharitonov, Xu,
  Mazar{\'e}, Karadayi, Liptchinsky, Collobert, Fuegen et~al.}]{kahn2020libri}
Jacob Kahn, Morgane Rivi{\`e}re, Weiyi Zheng, Evgeny Kharitonov, Qiantong Xu,
  Pierre-Emmanuel Mazar{\'e}, Julien Karadayi, Vitaliy Liptchinsky, Ronan
  Collobert, Christian Fuegen, et~al. 2020.
\newblock Libri-light: A benchmark for {ASR} with limited or no supervision.
\newblock In \emph{ICASSP 2020-2020 IEEE International Conference on Acoustics,
  Speech and Signal Processing (ICASSP)}, pages 7669--7673. IEEE.

\bibitem[{Kano et~al.(2021)Kano, Sakti, and Nakamura}]{kano2021transformer}
Takatomo Kano, Sakriani Sakti, and Satoshi Nakamura. 2021.
\newblock Transformer-based direct speech-to-speech translation with
  transcoder.
\newblock In \emph{2021 IEEE Spoken Language Technology Workshop (SLT)}, pages
  958--965. IEEE.

\bibitem[{Kong et~al.(2020)Kong, Kim, and Bae}]{kong2020hifi}
Jungil Kong, Jaehyeon Kim, and Jaekyoung Bae. 2020.
\newblock {H}i{F}i-{GAN}: Generative adversarial networks for efficient and
  high fidelity speech synthesis.
\newblock \emph{Advances in Neural Information Processing Systems}, 33.

\bibitem[{Kudo(2018)}]{kudo2018subword}
Taku Kudo. 2018.
\newblock Subword regularization: Improving neural network translation models
  with multiple subword candidates.
\newblock In \emph{Proceedings of the 56th Annual Meeting of the Association
  for Computational Linguistics (Volume 1: Long Papers)}, pages 66--75.

\bibitem[{Lakhotia et~al.(2021)Lakhotia, Kharitonov, Hsu, Adi, Polyak, Bolte,
  Nguyen, Copet, Baevski, Mohamed et~al.}]{lakhotia2021generative}
Kushal Lakhotia, Evgeny Kharitonov, Wei-Ning Hsu, Yossi Adi, Adam Polyak,
  Benjamin Bolte, Tu-Anh Nguyen, Jade Copet, Alexei Baevski, Adelrahman
  Mohamed, et~al. 2021.
\newblock Generative spoken language modeling from raw audio.
\newblock \emph{arXiv preprint arXiv:2102.01192}.

\bibitem[{Lavie et~al.(1997)Lavie, Waibel, Levin, Finke, Gates, Gavalda,
  Zeppenfeld, and Zhan}]{lavie1997janus}
Alon Lavie, Alex Waibel, Lori Levin, Michael Finke, Donna Gates, Marsal
  Gavalda, Torsten Zeppenfeld, and Puming Zhan. 1997.
\newblock {JANUS-III}: Speech-to-speech translation in multiple languages.
\newblock In \emph{1997 IEEE International Conference on Acoustics, Speech, and
  Signal Processing}, volume~1, pages 99--102. IEEE.

\bibitem[{Li et~al.(2019)Li, Liu, Liu, Zhao, and Liu}]{li2019neural}
Naihan Li, Shujie Liu, Yanqing Liu, Sheng Zhao, and Ming Liu. 2019.
\newblock Neural speech synthesis with transformer network.
\newblock In \emph{Proceedings of the AAAI Conference on Artificial
  Intelligence}, volume~33, pages 6706--6713.

\bibitem[{Li et~al.(2021)Li, Wang, Tang, Tran, Tang, Pino, Baevski, Conneau,
  and Auli}]{li2020multilingual}
Xian Li, Changhan Wang, Yun Tang, Chau Tran, Yuqing Tang, Juan Pino, Alexei
  Baevski, Alexis Conneau, and Michael Auli. 2021.
\newblock Multilingual speech translation from efficient finetuning of
  pretrained models.
\newblock In \emph{Proceedings of the 59th Annual Meeting of the Association
  for Computational Linguistics and the 11th International Joint Conference on
  Natural Language Processing (Volume 1: Long Papers)}, pages 827--838.

\bibitem[{Matusov et~al.(2005)Matusov, Kanthak, and
  Ney}]{matusov2005integration}
Evgeny Matusov, Stephan Kanthak, and Hermann Ney. 2005.
\newblock On the integration of speech recognition and statistical machine
  translation.
\newblock In \emph{Ninth European Conference on Speech Communication and
  Technology}.

\bibitem[{Nakamura et~al.(2006)Nakamura, Markov, Nakaiwa, Kikui, Kawai,
  Jitsuhiro, Zhang, Yamamoto, Sumita, and Yamamoto}]{nakamura2006atr}
Satoshi Nakamura, Konstantin Markov, Hiromi Nakaiwa, Gen-ichiro Kikui, Hisashi
  Kawai, Takatoshi Jitsuhiro, J-S Zhang, Hirofumi Yamamoto, Eiichiro Sumita,
  and Seiichi Yamamoto. 2006.
\newblock The {ATR} multilingual speech-to-speech translation system.
\newblock \emph{IEEE Transactions on Audio, Speech, and Language Processing},
  14(2):365--376.

\bibitem[{Ott et~al.(2019)Ott, Edunov, Baevski, Fan, Gross, Ng, Grangier, and
  Auli}]{ott2019fairseq}
Myle Ott, Sergey Edunov, Alexei Baevski, Angela Fan, Sam Gross, Nathan Ng,
  David Grangier, and Michael Auli. 2019.
\newblock fairseq: A fast, extensible toolkit for sequence modeling.
\newblock In \emph{Proceedings of NAACL-HLT 2019: Demonstrations}.

\bibitem[{Panayotov et~al.(2015)Panayotov, Chen, Povey, and
  Khudanpur}]{panayotov2015librispeech}
Vassil Panayotov, Guoguo Chen, Daniel Povey, and Sanjeev Khudanpur. 2015.
\newblock Librispeech: an {ASR} corpus based on public domain audio books.
\newblock In \emph{2015 IEEE international conference on acoustics, speech and
  signal processing (ICASSP)}, pages 5206--5210. IEEE.

\bibitem[{Park et~al.(2019)Park, Chan, Zhang, Chiu, Zoph, Cubuk, and
  Le}]{park2019specaugment}
Daniel~S Park, William Chan, Yu~Zhang, Chung-Cheng Chiu, Barret Zoph, Ekin~D
  Cubuk, and Quoc~V Le. 2019.
\newblock {S}pec{A}ugment: A simple data augmentation method for automatic
  speech recognition.
\newblock \emph{Proc. Interspeech 2019}, pages 2613--2617.

\bibitem[{Polyak et~al.(2021)Polyak, Adi, Copet, Kharitonov, Lakhotia, Hsu,
  Mohamed, and Dupoux}]{polyak2021speech}
Adam Polyak, Yossi Adi, Jade Copet, Eugene Kharitonov, Kushal Lakhotia,
  Wei-Ning Hsu, Abdelrahman Mohamed, and Emmanuel Dupoux. 2021.
\newblock Speech resynthesis from discrete disentangled self-supervised
  representations.
\newblock \emph{arXiv preprint arXiv:2104.00355}.

\bibitem[{Post(2018)}]{post2018call}
Matt Post. 2018.
\newblock A call for clarity in reporting {BLEU} scores.
\newblock In \emph{Proceedings of the Third Conference on Machine Translation:
  Research Papers}, pages 186--191.

\bibitem[{Post et~al.(2014)Post, Kumar, Lopez, Karakos, Callison-Burch, and
  Khudanpur}]{post2013improveddataset}
Matt Post, Gaurav Kumar, Adam Lopez, Damianos Karakos, Chris Callison-Burch,
  and Sanjeev Khudanpur. 2014.
\newblock {F}isher and {CALLHOME} {S}panish--{E}nglish speech translation.
\newblock LDC2014T23. Web Download. Philadelphia: Linguistic Data Consortium.

\bibitem[{Ren et~al.(2020)Ren, Hu, Tan, Qin, Zhao, Zhao, and
  Liu}]{ren2020fastspeech}
Yi~Ren, Chenxu Hu, Xu~Tan, Tao Qin, Sheng Zhao, Zhou Zhao, and Tie-Yan Liu.
  2020.
\newblock Fastspeech 2: Fast and high-quality end-to-end text to speech.
\newblock \emph{arXiv preprint arXiv:2006.04558}.

\bibitem[{Synnaeve et~al.(2019)Synnaeve, Xu, Kahn, Likhomanenko, Grave, Pratap,
  Sriram, Liptchinsky, and Collobert}]{synnaeve2019end}
Gabriel Synnaeve, Qiantong Xu, Jacob Kahn, Tatiana Likhomanenko, Edouard Grave,
  Vineel Pratap, Anuroop Sriram, Vitaliy Liptchinsky, and Ronan Collobert.
  2019.
\newblock End-to-end {ASR}: from supervised to semi-supervised learning with
  modern architectures.
\newblock \emph{arXiv preprint arXiv:1911.08460}.

\bibitem[{Tjandra et~al.(2019)Tjandra, Sakti, and Nakamura}]{tjandra2019speech}
Andros Tjandra, Sakriani Sakti, and Satoshi Nakamura. 2019.
\newblock Speech-to-speech translation between untranscribed unknown languages.
\newblock In \emph{2019 IEEE Automatic Speech Recognition and Understanding
  Workshop (ASRU)}, pages 593--600. IEEE.

\bibitem[{Vaswani et~al.(2017)Vaswani, Shazeer, Parmar, Uszkoreit, Jones,
  Gomez, Kaiser, and Polosukhin}]{vaswani2017attention}
Ashish Vaswani, Noam Shazeer, Niki Parmar, Jakob Uszkoreit, Llion Jones,
  Aidan~N Gomez, {\L}ukasz Kaiser, and Illia Polosukhin. 2017.
\newblock Attention is all you need.
\newblock In \emph{Advances in neural information processing systems}, pages
  5998--6008.

\bibitem[{Wang et~al.(2020{\natexlab{a}})Wang, Cho, and Gu}]{wang2020neural}
Changhan Wang, Kyunghyun Cho, and Jiatao Gu. 2020{\natexlab{a}}.
\newblock Neural machine translation with byte-level subwords.
\newblock In \emph{Proceedings of the AAAI Conference on Artificial
  Intelligence}, volume~34, pages 9154--9160.

\bibitem[{Wang et~al.(2021)Wang, Rivi{\`e}re, Lee, Wu, Talnikar, Haziza,
  Williamson, Pino, and Dupoux}]{wang2021voxpopuli}
Changhan Wang, Morgane Rivi{\`e}re, Ann Lee, Anne Wu, Chaitanya Talnikar,
  Daniel Haziza, Mary Williamson, Juan Pino, and Emmanuel Dupoux. 2021.
\newblock {V}ox{P}opuli: A large-scale multilingual speech corpus for
  representation learning, semi-supervised learning and interpretation.
\newblock \emph{arXiv preprint arXiv:2101.00390}.

\bibitem[{Wang et~al.(2020{\natexlab{b}})Wang, Tang, Ma, Wu, Okhonko, and
  Pino}]{wang2020fairseqs2t}
Changhan Wang, Yun Tang, Xutai Ma, Anne Wu, Dmytro Okhonko, and Juan Pino.
  2020{\natexlab{b}}.
\newblock fairseq {S2T}: Fast speech-to-text modeling with fairseq.
\newblock In \emph{Proceedings of the 2020 Conference of the Asian Chapter of
  the Association for Computational Linguistics (AACL): System Demonstrations}.

\bibitem[{Wang et~al.(2017)Wang, Skerry-Ryan, Stanton, Wu, Weiss, Jaitly, Yang,
  Xiao, Chen, Bengio et~al.}]{wang2017tacotron}
Yuxuan Wang, RJ~Skerry-Ryan, Daisy Stanton, Yonghui Wu, Ron~J Weiss, Navdeep
  Jaitly, Zongheng Yang, Ying Xiao, Zhifeng Chen, Samy Bengio, et~al. 2017.
\newblock Tacotron: Towards end-to-end speech synthesis.
\newblock \emph{Proc. Interspeech 2017}, pages 4006--4010.

\bibitem[{Weiss et~al.(2017)Weiss, Chorowski, Jaitly, Wu, and
  Chen}]{weiss2017sequence}
Ron~J Weiss, Jan Chorowski, Navdeep Jaitly, Yonghui Wu, and Zhifeng Chen. 2017.
\newblock Sequence-to-sequence models can directly translate foreign speech.
\newblock \emph{Proc. Interspeech 2017}, pages 2625--2629.

\bibitem[{Xu et~al.(2021)Xu, Baevski, Likhomanenko, Tomasello, Conneau,
  Collobert, Synnaeve, and Auli}]{xu2021self}
Qiantong Xu, Alexei Baevski, Tatiana Likhomanenko, Paden Tomasello, Alexis
  Conneau, Ronan Collobert, Gabriel Synnaeve, and Michael Auli. 2021.
\newblock Self-training and pre-training are complementary for speech
  recognition.
\newblock In \emph{ICASSP 2021-2021 IEEE International Conference on Acoustics,
  Speech and Signal Processing (ICASSP)}, pages 3030--3034. IEEE.

\bibitem[{Yang et~al.(2021)Yang, Chi, Chuang, Lai, Lakhotia, Lin, Liu, Shi,
  Chang, Lin et~al.}]{yang2021superb}
Shu-wen Yang, Po-Han Chi, Yung-Sung Chuang, Cheng-I~Jeff Lai, Kushal Lakhotia,
  Yist~Y Lin, Andy~T Liu, Jiatong Shi, Xuankai Chang, Guan-Ting Lin, et~al.
  2021.
\newblock {SUPERB}: Speech processing universal performance benchmark.
\newblock \emph{arXiv preprint arXiv:2105.01051}.

\bibitem[{Zhang et~al.(2020)Zhang, Tan, Ren, Qin, Zhang, and
  Liu}]{zhang2020uwspeech}
Chen Zhang, Xu~Tan, Yi~Ren, Tao Qin, Kejun Zhang, and Tie-Yan Liu. 2020.
\newblock {UWS}peech: Speech to speech translation for unwritten languages.
\newblock \emph{arXiv preprint arXiv:2006.07926}.

\end{thebibliography}
\bibliographystyle{acl_natbib}

\clearpage
\newpage 

\appendix
\section{Model training details}
\label{sec:model_config}
Table~\ref{tab:model_config} lists the hyper-parameters used in training direct S2ST models reported in Table~\ref{tab:written} and~\ref{tab:unwritten}. Model configurations are described in Sec.~\ref{sec:system_setup} and~\ref{sec:baseline}.

\begin{table}[hbt]
\centering
\resizebox{.99\linewidth}{!}{
\begin{tabular}{l|ccccc}
\hline
 ID & learning rate & dropout & max tokens per GPU & \# GPUs \\ 
\hline
4 & 0.001 & 0.3 & 80k & 16 \\
5 & 0.0005 & 0.1 & 80k & 16 \\
6-10 & 0.0005 & 0.1 & 20k & 4 \\
17 & 0.0005 & 0.1 & 20k & 4 \\
19 & 0.0001 & 0.1 & 20k & 4 \\
20 & 0.0005 & 0.1 & 20k & 4 \\
\hline
\end{tabular}
}
\caption{\label{tab:model_config} Training hyper-parameters for the direct S2ST models reported in Table~\ref{tab:written} and~\ref{tab:unwritten}.
}
\end{table}

\section{Examples of model output}
\label{sec:text_example}
Table~\ref{tab:text_exp} shows examples of the ASR decoded text on the speech output and the text output from CTC decoding. 
As shown in Table~\ref{tab:text_exp}, the generated speech and the CTC decoded text are consistent with each other, while the auxiliary task may generate inconsistent text output due to a separate attention module. The small mismatch between the model's speech output and the CTC decoded text is due to a combination of ASR errors and misspelling from CTC decoding.

\begin{table*}[ht!]
\centering
\begin{tabular}{l|l}
\hline
\textit{human} & i've been living here for twenty six years \\
ASR & i've been living here for twenty six years \\
CTC & i've been living here for \textbf{twentysix} years \\
\textit{tc} & \textbf{i'm twentysix years} living here \\
\textit{ref} & i've been living here twenty six years \\
\hline
\textit{human} & but but i don't go to puerto rico because i have two kids here \\
ASR & but \textbf{$\ast$} i don't go to \textbf{porto} rico because i have two kids here \\
CTC & but but i don't go to puerto rico because i have two kids here \\
\textit{tc} & but but \textbf{i'm not going} to \textbf{live there} puerto rico because i have \textbf{thousand} two kids here \\
\textit{ref} & but but i'm not going to live there in puerto rico because i have my two children here \\
\hline
\textit{human} & oh yeah that was that what do you think about interracial marriage \\
ASR & oh \textbf{yes} that was that what do you think about \textbf{inter ratial} marriage \\
CTC & oh yeah that was that what do you think about interracial marriage \\
\textit{tc} & oh yeah \textbf{that's $\ast$ $\ast$ $\ast$ $\ast$ $\ast$ the} interracial marriage \\
\textit{ref} & oh yeah that was that was today's subject so what do you think about interracial marriage \\
\hline
\end{tabular}
\caption{\label{tab:text_exp} Examples of output from our best model under the written language setup, ``S2UT \textit{reduced} + CTC (w/ \textit{sc, tc})``. We compare text from
\begin{enumerate*}[label=(\arabic*)]
  \item \textit{human}: human transcription of the generated audio,
  \item ASR: ASR decoded text on the generated audio,
  \item CTC: the model's text output from CTC decoding,
  \item \textit{tc}: output from the model's auxiliary task trained with target characters as targets, and
  \item \textit{ref}: ground truth reference translation.
\end{enumerate*}
The differences with respect to \textit{human} are highlighted in bold for the text from ASR, CTC and \textit{tc}, and $\ast$ denotes word deletion.
}
\end{table*}

\section{Significance test}
\label{sec:sig_test}
Table~\ref{tab:sig_test} shows the \textit{p}-values from paired significance tests between the nine systems (ID 2-10) in Table~\ref{tab:written}. We conduct the tests with the paired bootstrap resampling method supported in the \textsc{sacreBLEU} tool~\citep{post2018call}.

\begin{table*}[ht!]
\centering
\begin{tabular}{c|cccccccc}
ID & 2 & 3 & 4 & 5 & 6 & 7 & 8 & 9 \\ 
\hline
3 & 0.0010$^{\ast}$ & - & -  & -  & -  & -  & -  & - \\
4 & 0.0010$^{\ast}$ & 0.0010$^{\ast}$ & -  & -  & -  & -  & -  & - \\
5 & 0.0010$^{\ast}$ & 0.0010$^{\ast}$ & 0.0010$^{\ast}$ & -  & -  & -  & -  & -  \\
6 & 0.0010$^{\ast}$ & 0.0010$^{\ast}$ & 0.0010$^{\ast}$ & 0.0150$^{\ast}$ & - & -  & -  & - \\
7 & 0.0010$^{\ast}$ & 0.0010$^{\ast}$ & 0.0010$^{\ast}$ & 0.0060$^{\ast}$ & 0.2218 & - & -  & -  \\
8 & 0.0010$^{\ast}$ & 0.0010$^{\ast}$ & 0.0010$^{\ast}$ & 0.0090$^{\ast}$ & 0.2018 & 0.4206 & - & - \\
9 & 0.0010$^{\ast}$ & 0.0010$^{\ast}$ & 0.0010$^{\ast}$ & 0.0010$^{\ast}$ & 0.0010$^{\ast}$ & 0.0010$^{\ast}$ & 0.0010$^{\ast}$ & - \\
10 & 0.0010$^{\ast}$ & 0.1798 & 0.0010$^{\ast}$ & 0.0010$^{\ast}$ & 0.0010$^{\ast}$ & 0.0010$^{\ast}$ & 0.0010$^{\ast}$ & 0.0010$^{\ast}$ \\
\end{tabular}
\caption{\label{tab:sig_test} \textit{p}-values from paired significance tests between nine systems (ID 2-10) in Table~\ref{tab:written} on the Fisher test set. \textit{p}-values $< 0.05$ are marked with ``$\ast$''.
}
\end{table*}

\end{document}